\documentclass[a4paper,conference]{IEEEtran}
\IEEEoverridecommandlockouts
\usepackage{cite}
\usepackage{amsmath,amssymb,amsfonts}
\usepackage{algorithmic}
\usepackage{graphicx}
\usepackage{textcomp}
\usepackage{xcolor}
\usepackage{import}
\usepackage{hyperref}
\usepackage{multirow}
\usepackage{amsmath}
\usepackage{bm}
\usepackage{array}
\newcolumntype{P}[1]{>{\centering\arraybackslash}m{#1}}
\hypersetup{colorlinks=false,breaklinks=true}
\def\BibTeX{{\rm B\kern-.05em{\sc i\kern-.025em b}\kern-.08em
    T\kern-.1667em\lower.7ex\hbox{E}\kern-.125emX}}
\begin{document}

\title{Improving Word Recognition using Multiple Hypotheses and Deep Embeddings}

\author{\IEEEauthorblockN{Siddhant Bansal}
\IEEEauthorblockA{\textit{CVIT, IIIT, Hyderabad, India} \\
siddhant.bansal@students.iiit.ac.in}
\and
\IEEEauthorblockN{Praveen Krishnan}
\IEEEauthorblockA{\textit{CVIT, IIIT, Hyderabad, India} \\
praveen.krishnan@research.iiit.ac.in}
\and
\IEEEauthorblockN{C.V. Jawahar}
\IEEEauthorblockA{\textit{CVIT, IIIT, Hyderabad, India} \\
jawahar@iiit.ac.in}
}

\maketitle

\begin{abstract}
% We propose to improve word recognition by fusing deep word image embeddings to the recognition pipeline using learning-based approaches.
% We propose to improve word recognition by fusing it with word image embeddings using learning-based approaches.
We propose a novel scheme for improving the word recognition accuracy using word image embeddings.
% We propose to fuse recognition-based and word spotting approaches for improving word recognition (of segmented words) using learning-based methods.
% We propose to fuse recognition-based and recognition-free approaches for word recognition using learning-based methods. 
% For this purpose, results obtained using a text recognizer and deep embeddings (generated using an End2End network) are fused.
% For this purpose, textual transcriptions obtained using a text recognizer and deep word image embeddings generated using an End2End network, are fused.
We use a trained text recognizer, which can predict multiple text hypothesis for a given word image.  Our fusion scheme improves the recognition process by utilizing the word image and text embeddings obtained from a trained word image embedding network.
We propose EmbedNet, which is trained using a triplet loss for learning a suitable embedding space where the embedding of the word image lies closer to the embedding of the corresponding text transcription.
The updated embedding space thus helps in choosing the correct prediction with higher confidence.
% This updated embedding space helps in choosing the correct prediction with higher confidence.
To further improve the accuracy, we propose a plug-and-play module called Confidence based Accuracy Booster (\textsc{cab}). 
The \textsc{cab} module takes in the confidence scores obtained from the text recognizer and Euclidean distances between the embeddings to generate an updated distance vector.
% It takes in the confidence scores obtained from the text recognizer and Euclidean distances between the embeddings and generates an updated distance vector.
The updated distance vector has lower distance values for the correct words and higher distance values for the incorrect words.
We rigorously evaluate our proposed method systematically on a collection of books in the Hindi language.
Our method achieves an absolute improvement of around 10\% in terms of word recognition accuracy.
\end{abstract}

\begin{IEEEkeywords}
Word recognition, word image embedding, EmbedNet
\end{IEEEkeywords}

\section{Introduction} \label{introduction}
% Paragraph on how important word recognition is.
The task of word recognition involves converting the text in an image to a machine-readable format.
Word recognition is an important use case of computer vision that finds various applications in digitizing old books, making self-driving cars understand signboard instructions, and creating assistive applications for people with special needs.
All these tasks rely on accurate word recognition that is robust to extreme variations in lighting conditions, fonts, sizes and overall typography.
To ensure the availability of the word recognizer to a broader audience, it should also be able to function for various languages and have low computational costs.

\begin{figure}[!tphb]
    \centering
    \includegraphics[width=0.48\textwidth]{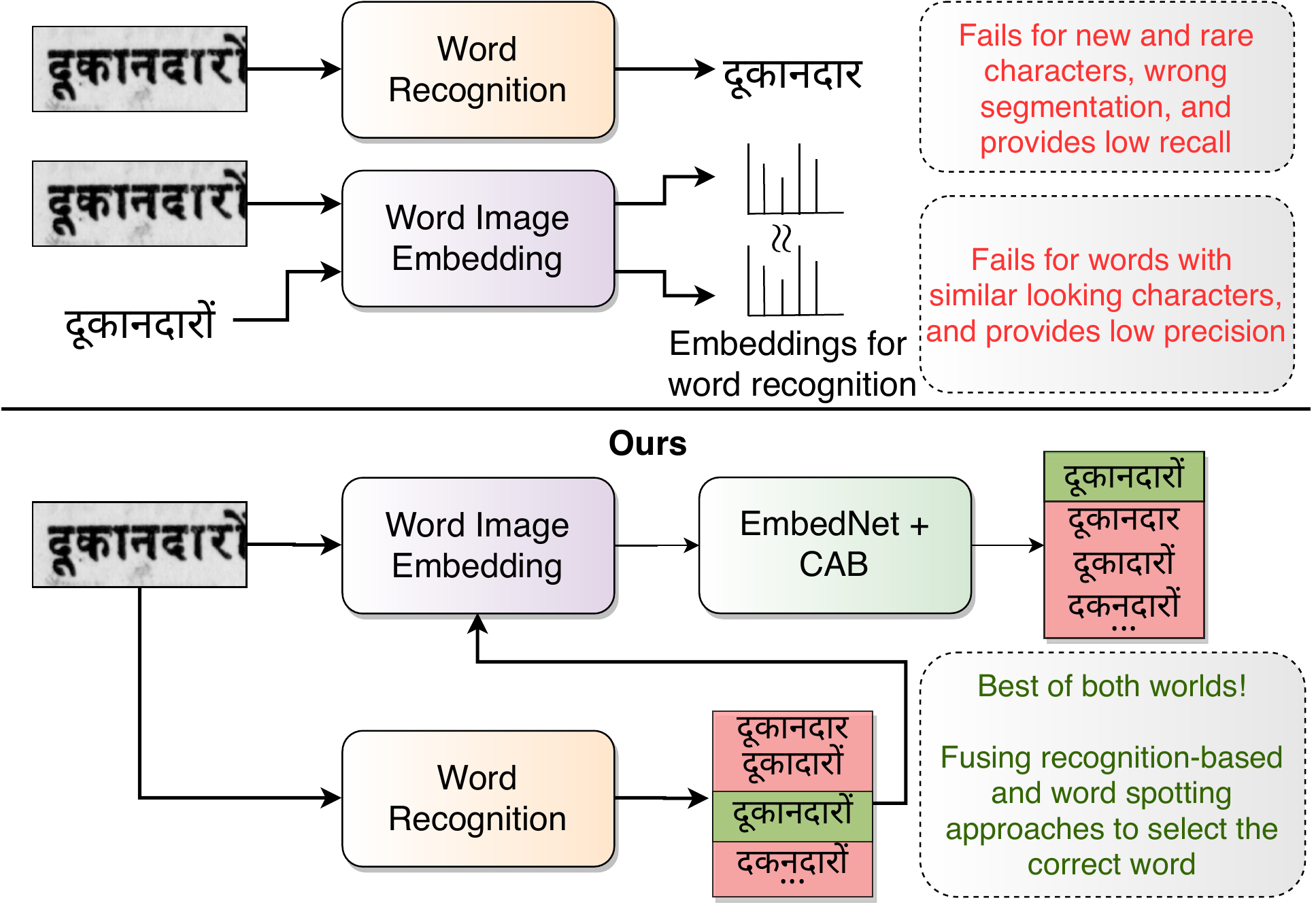}
    \caption{Word recognition based methods fail to perform when they encounter new and rare characters, wrong word image segmentation, and provide low recall. However, these methods excel in differentiating between visually similar characters, whereas it is vice-versa for methods using word image embeddings.
    % Recognition-based methods fail to perform when they encounter new and rare characters, but excel in differentiating between visually similar characters, whereas it is vice-versa for recognition-free methods.
    We aim to use the methods proposed in this work - EmbedNet and \textsc{cab}, for exploring the complementary properties of these methods.
    % We aim to explore the complementary properties of these methods and fuse them to improve word recognition.
    Diagram best viewed in color.
    }
    \label{fig:idea}
\end{figure}

% Paragraph on challenges faced while working with the Hindi language.
In this work, we focus on improving word recognition for the Hindi language, which is agglutinative and inflectional.
% Additionally, words in Hindi are usually longer, and amass to form an immense vocabulary.
Hindi contains $11$ vowels and $33$ consonants, and a horizontal line runs across the words, which is referred to as \textit{Shirorekha}.
If a consonant is followed by a vowel, the shape of the consonant is modified.
Such characters are referred to as vowel modifiers.
A compound character is formed when a consonant follows one or more consonants.
Due to these modifiers and compound characters, the number of distinct shapes in Hindi is far more than that of the Latin scripts~\cite{why_hindi}.
This makes word recognition for Hindi difficult, and hence, it is necessary to devise more intricate techniques which improve the word recognition accuracy for the Hindi language.

Traditionally, word recognition methods fall under two major categories: (a) methods directly converting a word image to its textual transcription~\cite{CRNN_begins,Chen,Sun,Garain,Adak}, and (b) methods converting word images to embeddings and then performing recognition using these embeddings~\cite{Kris,Sudholt,Wilkinson,Sudhlt}.
In this work, we will refer to methods in category (a) as word recognition and methods in category (b) as word image embeddings, respectively.
We assume that the word images are already segmented.
A word recognition based method aims at directly converting the word image to its corresponding textual transcription.
% Despite the wide availability of high-grade open-source \textsc{ocr} engines \cite{Tesseract}, using them with degraded images from historical documents proves to be ineffective, and limits the usage of \textsc{ocr}.
Despite the wide availability of high-grade open-source \textsc{ocr} engines \cite{Tesseract}, using them with degraded images from historical documents is difficult.
On the other hand, word image embedding methods focus on converting the word image and the corresponding text to a holistic representation where word image and its corresponding text lie closer to one another.
After the projection of these images and texts to a learned representation space, they can be compared using an appropriate distance metric and perform recognition restricted to a lexicon~\cite{Kris}.
% After projecting the images and text to the learned representation space, one can compare these using an appropriate distance metric and perform recognition restricted to a lexicon~\cite{Kris}.

% Word recognition methods (\textsc{ocr}) perform reasonably well when the text present in the image comprises of characters which have been adequately provided in the training data.
Word recognition methods (\textsc{ocr}) perform reasonably well when the text present in the image is reasonably clean.
% Word recognition methods (\textsc{ocr}) perform adequately and generate excellent results when the text present in the image comprises of characters which have enough occurrences in the training data.
However, if the \textsc{ocr} encounters a rare character or an image with higher degradation, it struggles to generate the correct prediction.
In such cases, word image embedding methods prove to be much useful.
The reason is that they do not identify each character but instead, focus on converting the word image to an embedding/representation where words with (visually) similar characters, lie closer in the embedding space, resulting in better predictions in challenging situations.
% This results in better predictions in challenging situations.
% For such cases, word image embedding methods prove to be quite useful as they do not identify each character; they focus on converting the word image to an embedding/representation where words with similar characters (visually) lie closer in the embedding space.
However, word image embedding methods find it difficult to distinguish between two different words with an approximately similar set of characters, a task at which word recognition based approaches excel.
% However, word image embedding methods find difficulty in differentiating between two different words with an approximately similar set of characters, the task at which word recognition based approaches excel.
Also, word recognition methods provide high recall, whereas, word image embeddings methods, provide high precision~\cite{prec_recall}.
% Apart from the aforementioned differences between word recognition and word image embeddings approaches, another difference of interest is that word recognition methods provide high recall, whereas, word image embeddings methods provide high precision~\cite{prec_recall}.
% Also, as shown in~\cite{word_seg}, inaccurate word segmentation results in poor performance when using word recognition methods.
Inaccurate word segmentation degrades performance in word recognition based methods~\cite{word_seg}. 
Inaccurate segmentation, however, does not hinder the performance of word image embedding as a slightly degraded word (due to cut) still lies closer to its textual transcription's embedding.
% However, inaccurate word segmentation does not hinder the performance of word image embedding as the images are converted to a holistic representation where a slightly degraded word (due to cut) still lies closer to its textual transcription's embedding.
% Fig.~\ref{fig:idea} summarises the drawbacks of both the methods and shows how we use these complementary properties to improve word recognition.
Fig.~\ref{fig:idea}, we show how we propose to use the complementary properties of both methods for improving word recognition.
% Fig.~\ref{fig:idea} summarises the drawbacks of word recognition and word image embeddings based methods and shows how we fuse both of these methods to improve word recognition.
% Word retrieval systems using word spotting are known to have a higher recall, whereas, text recognition-based systems provide higher precision~\cite{prec_recall}.

Designing a pipeline that can exploit the complementary properties provided by word recognition and word image embedding methods can further enhance word recognition. In our previous work~\cite{sid_das}, we propose to use the complementary information of both the methods to create a more reliable and robust word recognition algorithm.
% Designing a pipeline that can exploit the complementary properties provided by recognition-based and recognition-free methods can further enhance word recognition. Siddhant et al. \cite{sid_das} propose to use the complementary information of both the methods to create a more reliable and robust word recognition algorithm.
We propose to fuse multiple hypotheses generated by the word recognizer with the embeddings generated from the End2End network (`\textit{E2E}')~\cite{e2e} for making use of the complementary information.
Using the beam search decoding algorithm, we produce multiple ($K$) predictions for a word image from a \textsc{ctc}~\cite{ctc} based word recognizer, where $K$ is the number of predictions generated for a word image.
We show that as the value of $K$ increases, the word recognition accuracy increases.
Even though we proposed multiple rule-based methods for using this information and improving word recognition accuracy, we do not explore the learning-based techniques in~\cite{sid_das}.

In this work, we improve upon the methods presented in~\cite{sid_das} and propose EmbedNet and a novel plug-and-play module called Confidence based Accuracy Booster (\textsc{cab}) for improving word recognition.
Fig.~\ref{fig:entire_flow} presents the flowchart of the entire process which includes EmbedNet, \textsc{cab}, and their roles in the word recognition pipeline.
Here, EmbedNet attempts to learn an updated Euclidean space where the embeddings of the word image and its correct textual transcription lie closer together, while the incorrect ones lie farther away.
The \textsc{cab} boosts the word recognition accuracy by using the updated representation made available by EmbedNet.
For accelerating future research, we release the code and models used in this work on our webpage\footnote{\href{http://cvit.iiit.ac.in/research/projects/cvit-projects/word-recognition}{http://cvit.iiit.ac.in/research/projects/cvit-projects/word-recognition}}.

\section{Related Works} \label{related_works}
In this work, we are interested in devising deep learning methods for fusing the existing methods in the word recognition, and word image embedding realms also referred to as text recognition and word spotting, respectively.
% In this work, we are interested in devising deep learning methods for fusing the existing methods in the recognition-based and recognition-free realms, which are also referred to as text recognition and word spotting, respectively.
This section explores the previous work done in these domains.

\subsection{Text Recognition}
A typical text recognizer involves a feature extractor for the input image containing text, and a sequential encoder for learning the temporal information.
Modern text recognizers use Convolutional Neural Networks (\textsc{cnn}) as a feature extractor and a Recurrent Neural Network (\textsc{rnn}) as a sequential encoder, which helps in modeling the text recognition problem as a Seq2Seq problem.
Architectures using both \textsc{cnn} and \textsc{rnn} for this purpose are called Convolutional Recurrent Neural Network (\textsc{crnn})~\cite{CRNN_begins}.
Previous works have used a wide range of recurrent networks for encoding the temporal information.
Adak et al.~\cite{Adak} perform sequential classification using a \textsc{rnn}, whereas,~\cite{Garain} use Bi-directional Long-Short Term Memory (\textsc{blstm}) network for sequential classification using Connectionist Temporal Classification (\textsc{ctc}) loss~\cite{ctc}.
Sun et al.~\cite{Sun} propose to use multi-directional \textsc{lstm} as the recurrent unit, whereas, Chen et al.~\cite{Chen} use Separable Multi-Dimensional Long Short-Term Memory for the same.
These methods attempt to address word recognition by undertaking the task of directly converting an input document to a machine-readable text.

\subsection{Word Spotting}
Word spotting~\cite{manmatha96} is an alternative for word recognition where we formulate a matching task. 
The fundamental problem in word spotting is about learning an appropriate feature representation for word images which is suitable for matching within the collections of document images. 
In this paper, we consider the word level segmentation to be available a-priori, and thereby limit our discussion on works which are in the domain of segmentation-based word spotting.
An initial method~\cite{Rath} represents the word image using profile features and then uses different distance metrics for comparing them.
Other works use handcrafted features~\cite{Balasubramanian}, and Bag of Visual Words~\cite{Shekhar} for word spotting. 
Most of these early representations were learned in an unsupervised way. 
The later methods drifted towards learning in a supervised setting and presented robust representation schemes. 
One of the classical methods in this space is from Almazan et al.~\cite{AlmazanPAMI14} which introduced an attributes framework referred to as Pyramidal Histogram of Characters ({\sc phoc}) for representing both images and text. 
More recently, various deep learning based approaches~\cite{Jaderberg_2,Jaderberg_3} have improved word spotting.
VGGNet~\cite{VGG} was adopted by Poznanski et al.~\cite{Poznanski} for recognising \textsc{phoc} attributes.
Many other methods in the word spotting domain successfully explored using \textsc{phoc} as the embedding spaces through different \textsc{cnn} architectures~\cite{Kris,Sudholt,Wilkinson,Sudhlt}.
In the Indian language document community, methods like~\cite{Bhardwaj,Chaudhury,Shekhar} attempt word spotting methods on Indian texts.
% None of the methods mentioned attempts to fuse the recognition-based and recognition-free approach, whereas, in this work, 

In this work, we propose to combine a \textsc{cnn-rnn} architecture proposed in~\cite{stn_crnn} with the embeddings generated from the \textit{E2E} proposed in~\cite{e2e} using learning-based methods.
By combining two different approaches, we aim at assimilating the best attributes of both the methods.

% In our earlier work~\cite{sid_das}, we attempt to fuse word recognition and word image embedding approaches using $K$ hypotheses from a text recognizer and deep embeddings generated using \textit{E2E} \cite{e2e}.
% % The work \cite{sid_das} which closely resembles this work attempts to fuse recognition-based and recognition-free approaches using $K$ hypotheses from a text recognizer and deep embeddings generated using \textit{E2E} \cite{e2e}.
% However, it was limited to rule-based methods. In this work, we propose EmbedNet, which takes in the embeddings from the \textit{E2E} network and updates them.
% This helps in improving the word recognition accuracy.
% To further boost the word recognition accuracy, we propose Confidence based Accuracy Booster (\textsc{cab}).
% \textsc{cab} takes in the confidence scores from the text recognizer and Euclidean distances between the embeddings to generate an updated distance vector.
% %Authors in \cite{sid_das} also propose to use the confidence score generated from the text recognizer by merely adding it with the distances generated using the deep embeddings.
% %Using this method, they were able to improve the word recognition accuracy up to a certain value of $K$, after which it starts to decrease.
% In this work, we also propose Confidence based Accuracy Booster (\textsc{cab}), which overcomes this drawback and achieves a consistent accuracy as the value of $K$ increases.

\section{Methods for improving word recognition} \label{methodology}
In this section, we elaborate on the proposed method using EmbedNet and \textsc{cab} .
This section is divided as follows, 
Section \ref{word_recognition} and \ref{word_spotting} brief about \textsc{crnn}~\cite{stn_crnn} and the End2End network~\cite{e2e}, respectively.
% Section \ref{word_recognition} briefs the \textsc{crnn}~\cite{stn_crnn} architecture used as a text recogniser.
% Section \ref{word_spotting} briefs the End2End network~\cite{e2e} used for word spotting.
In Section \ref{EmbedNet_intro}, we motivate EmbedNet and Section \ref{cab} proposes a novel Confidence based Accuracy Booster (\textsc{cab}), a plug-and-play module for boosting the word recognition accuracy.
% As done in \cite{sid_das}, we use the following methods for word recognition and word spotting. 
% [maybe we can add this] More details are presented in their original work.

\subsection{Word Recognition} \label{word_recognition}
We use a standard \textsc{cnn-rnn} ({\sc crnn}) hybrid architecture that was proposed in \cite{stn_crnn}.
The network converts the textual contents of an image to textual transcriptions.
Fig. \ref{fig:entire_flow} shows the architecture of the \textsc{crnn} architecture used in our work. 
It consists of a spatial transformer layer
(\textsc{stn}) followed by the residual convolutional blocks. These blocks are responsible for learning a sequence of feature maps using ResNet18 \cite{resnet18}.
These feature sequences serve as an input into a stacked bi-directional long short-term memory (\textsc{blstm}) network.
% It takes in every frame from the feature sequence as an input and generates the probability distribution over the class labels.
% Connectionist Temporal Classification (\textsc{ctc}) \cite{ctc} is used for decoding the target label sequence over all the frames using the probabilities generated by the \textsc{blstm} network.
Connectionist Temporal Classification (\textsc{ctc}) \cite{ctc} is then used for decoding the target label sequence over all the frames.

\subsection{Word Spotting} \label{word_spotting}
Fig. \ref{fig:entire_flow} shows the End2End network proposed in \cite{e2e} which learns the textual and visual word image embeddings.
The network consists of two major input streams: real and label.
In the real stream, ResNet34 contributes by generating the features for the real images.
The label stream further gets divided into: (a) synthetic image stream and (b) text stream.
Synthetic image's feature extraction takes place with the help of a shallow \textsc{cnn} architecture, while, generation of textual features happens using a \textsc{phoc} extractor.
The features generated are then appended and treated as a conditional label and merged using a fully connected network.
The features generated from these streams are appended and merged using a fully connected network which preserves information from both modalities.
This fully connected network preserves information from both modalities.
After this, the embedding layer projects the embeddings from the real and label stream to a common subspace.

\subsection{EmbedNet} \label{EmbedNet_intro}
\begin{figure*}[!htpb]
    \centering
    \includegraphics[width=\textwidth]{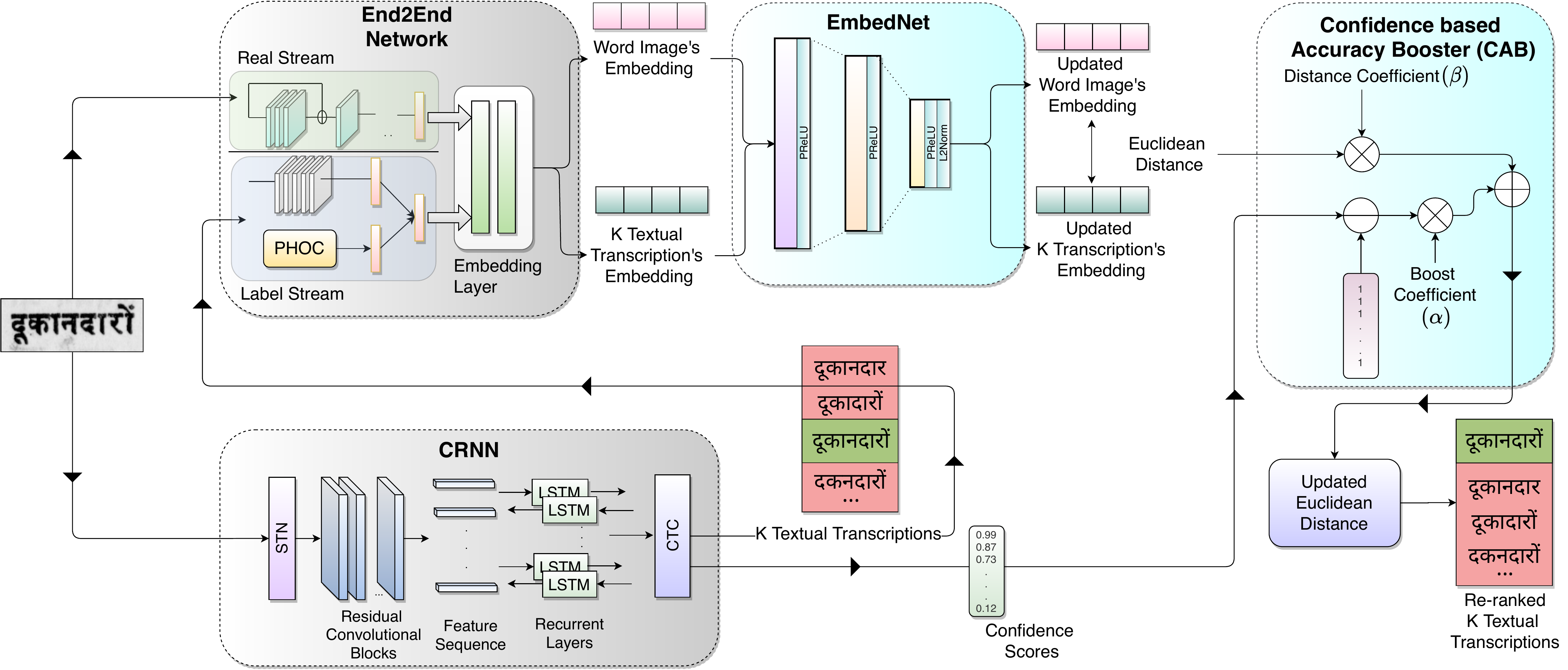}
    \caption{For generating the textual transcription, we pass the word image through the \textsc{crnn} \cite{stn_crnn} and the End2End network (`\textit{E2E}') \cite{e2e}, simultaneously.
    The \textsc{crnn} generates multiple ($K$) textual transcriptions for the input image, whereas the \textit{E2E} network generates the word image's embedding.
    The $K$ textual transcriptions generated by the \textsc{crnn} are passed through the \textit{E2E} network to generate their embeddings.
    We pass these embeddings through the EmbedNet proposed in this work.
    The EmbedNet projects the input embedding to an updated Euclidean space, using which we get updated word image embedding and $K$ transcriptions' embedding.
    We calculate the Euclidean distance between the input embedding and each of the $K$ textual transcriptions.
    We then pass the distance values through the novel Confidence based Accuracy Booster (\textsc{cab}), which uses them and the confidence scores from the \textsc{crnn} to generate an updated list of Euclidean distance, which helps in selecting the correct prediction.
    Diagram best viewed in color.
    }
    \label{fig:entire_flow}
\end{figure*}

In this work, we propose EmbedNet for projecting the embeddings to an updated embedding space and \textsc{cab} for boosting the word recognition accuracy.
We use a set of $n$ word images for which we want to get the textual transcriptions.
As shown in Fig. \ref{fig:entire_flow}, these $n$ images are passed through the real stream of \textit{E2E} to generate embeddings represented by $\phi_i \; \forall i \in 1, \dots, n$.
% Beam search decoding algorithm with \textsc{crnn} produces $K$ predictions for each of the $n$ word images.
$K$ predictions generated for each of the $n$ word image are converted to embeddings symbolized by $\psi_i^j \; \forall i \in 1, \dots, n; \forall j \in 1,\dots, K$ using the \textit{E2E}'s label stream.
We set the value of $K$ equal to $20$ throughout the paper, unless otherwise specified.

\begin{figure}[!thpb]
    % \centerline{\includegraphics[width=0.5\textwidth]{./images/EmbedNet.png}}
    \centering
    \includegraphics[width=0.48\textwidth]{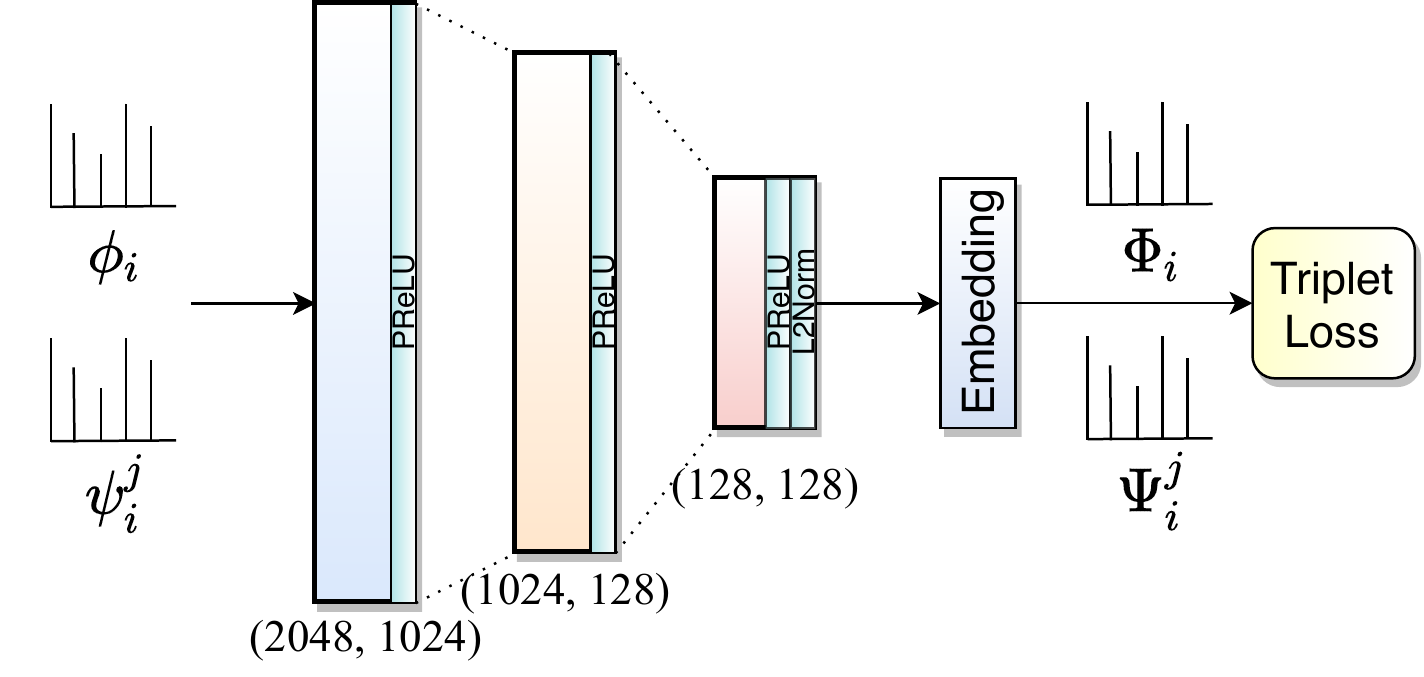}
    \caption{An EmbedNet, during training, takes in a word image's embedding ($\phi_i^a$), correct text's embedding ($\psi^{+}_k$) and incorrect text's embedding ($\psi^{\mbox{-}}_l$) one at a time.
    Corresponding output embeddings are passed through the triplet loss for training.
    Once trained, it takes in $\phi_i$ and $\psi_i^j$ and generates $\Phi_i$ and $\Psi_i^j$, respectively.
    % While determining the best EmbedNet architecture, we can omit the blocks with a dashed boundary.
    Tuples underneath the blocks represent the input and output size of the corresponding block.
    See text for notation.}
    \label{fig:EmbedNet}
\end{figure}{}

% EmbedNet improves over the \textsc{mlp} by learning a compact Euclidean space where the correct $\psi_i^j$ lies closer to $\phi_i$, and incorrect $\psi_i^j$ lies farther away from $\phi_i$.
% EmbedNet learns a compact Euclidean space where the correct $\psi_i^j$ lies closer to $\phi_i$, and incorrect $\psi_i^j$ lies farther away from $\phi_i$.
We generate $\Phi_i$ and $\Psi_i^j$ by providing $\phi_i$ and $\psi_i^j$ as inputs to the EmbedNet, respectively.
% EmbedNet takes in $\phi_i$ and $\psi_i^j$ and generates $\Phi_i$ and $\Psi_i^j$, respectively.
Fig. \ref{fig:EmbedNet} shows the EmbedNet architecture; it projects the embeddings from $\Re^{2048}$ to $\Re^{128}$ using a $2048$ dimensional linear input layer and $128$ dimensional linear output layer and a hidden layer in between.
% Table \ref{tab:EmbedNet_comparison} summarises the ablation study performed for finalizing the number of hidden layers.
% From the study, we conclude that the architecture with one hidden layer performs the best.
% ; now onwards, we refer to this architecture as EmbedNet.
We add PReLU activation function after each layer; it helps in introducing non-linearity to the model.
L2 normalization is performed on the final layer's output to project the embedding on a $128$-dimensional hyper-sphere.
We train the EmbedNet for $200$ epochs with early stopping acting as a regularizer and use the Adam optimizer with a constant learning rate of $0.0001$.

% We achieve a compact Euclidean space by training the EmbedNet using triplet pairs originally proposed in \cite{facenet}.
% The triplet pairs consist of an anchor, a positive and a negative embedding are generated using $\phi_i$ and $\psi_i^j$.
% Here, the anchor is the embedding of the word image for which we want to generate the textual transcription.
% It is denoted as $\phi_i^a \; \forall i = 1, \dots, n$.
% Positive and negative are $c$ correct and $w$ incorrect textual transcription's embeddings, respectively.
% Here, $c + w \leq n \times K$.
% Positive embeddings are denoted as $\psi^{+}_k \; \forall k \in 1, \dots, c$, and negative embeddings are denoted as $\psi^{\mbox{-}}_l \; \forall l \in 1, \dots, w$.

Let EmbedNet be a function $f_{en}$ defined as $f_{en}(\phi_i, \psi_i^j) = \Phi_i, \Psi_i^j$; it learns a compact Euclidean space where the correct $\Psi_i^j$ lies closer to $\Phi_i$, and incorrect $\Psi_i^j$ lies farther away from $\Phi_i$.
We achieve the compact Euclidean space by training the EmbedNet using the triplet pairs as originally proposed in \cite{facenet}.
The pairs constitute of three different embeddings; the first one is the embedding of the word image from the train set for which we want to generate the textual transcription; we refer to them as the anchor denoted as $\phi_i^a \; \forall i = 1, \dots, n$.
The second one is the embedding of $c$ words for which we have correct textual transcription; we call them positive denoted as $\psi^{+}_k \; \forall k \in 1, \dots, c$.
The third one is the embedding of $w$ words with incorrect textual transcription; we refer to them as negative denoted as $\psi^{\mbox{-}}_l \; \forall l \in 1, \dots, w$.
We sample the anchor from $\phi_i$, and $\psi_i^j$ is sampled for generating positive and negative.

The triplets are further classified into:
% here $\gamma$ is the margin enforced between positive and negative pairs.
\paragraph{Hard Negatives} Equation \ref{eq1} shows the condition for hard negatives; here, the Euclidean distance between the anchor and positive is greater than the distance between the anchor and negative.
Due to this, they contribute the most while training the EmbedNet.
\begin{equation}
    || \phi_i^a - \psi^{+}_k ||^2_2 > || \phi_i^a - \psi^{\mbox{-}}_l||^2_2
    \label{eq1}
\end{equation}
% In equation \ref{eq1}, the Euclidean distance between the anchor and positive is greater than the distance between the anchor and negative.
% Such pairs are called \textbf{hard negatives} and contribute the most when training the EmbedNet.
\paragraph{Semi-hard Negatives} Equation \ref{eq2} defines the condition for semi-hard negatives; it relies on the margin ($\gamma$). Here the Euclidean distance between the anchor and negative is less than the distance between the anchor and positive $+\; \gamma$ but higher than the Euclidean distance between the anchor and positive.
\begin{equation}
    || \phi_i^a - \psi^{+}_k ||^2_2 < || \phi_i^a - \psi^{\mbox{-}}_l||^2_2 < || \phi_i^a - \psi^{+}_k ||^2_2 + \gamma
    \label{eq2}
\end{equation}

\paragraph{Easy Negatives} Equation \ref{eq3} shows the condition for easy negatives; here, the Euclidean distance between the anchor and positive $+\; \gamma$ is less than the distance between the anchor and negative.
\begin{equation}
    || \phi_i^a - \psi^{+}_k ||^2_2 + \gamma < || \phi_i^a - \psi^{\mbox{-}}_l||^2_2
    \label{eq3}
\end{equation}
% Equation \ref{eq2} defines \textbf{semi-hard negatives} based on the margin ($\gamma$), here the Euclidean distance between the anchor and negative is less as compared to the distance between the anchor and positive $+\; \gamma$, but, higher than the Euclidean distance between the anchor and positive.
% Equation \ref{eq3} defines the \textbf{easy negatives}, here the Euclidean distance between the anchor and positive $+\; \gamma$ is less than the distance between the anchor and negative.

Easy negatives do not contribute while training the EmbedNet as the condition of Euclidean distance between the anchor and positive example being less than the distance between the anchor, and negative is already satisfied.
We train the EmbedNet using the Triplet loss, it is defined as:
\begin{equation}
    L(\phi_i^a, \psi_k^{+}, \psi^{\mbox{-}}_l) = max(||\phi_i^a - \psi_k^{+}||_2^2 - ||\phi_i^a - \psi^{\mbox{-}}_l||_2^2 + \gamma, 0)
\end{equation}
here $\phi_i^a, \psi_k^{+}$ and $\psi^{\mbox{-}}_l$ are anchor, positive and negative embeddings respectively and $\gamma$ is the margin.

% [TODO] ADD DIAGRAM EXPLAINING THE TRIPLETS PROCESS

The triplet pairs are updated after every epoch.
For updating, we pass $\phi_i$ and $\psi_i^j$ through the EmbedNet, identify anchors, positives, and negatives.
They are then further divided into hard negatives, semi-hard negatives, and easy negatives using equations \ref{eq1}, \ref{eq2}, and \ref{eq3}, respectively.

% As the training progresses, the EmbedNet starts to over-fit on the initial triplets generated, to avoid it, the triplet pairs are updated regularly.
% For updating, we pass all the triplets through the EmbedNet and again divide triplets into hard negatives, semi-hard negatives, and easy negatives using equations \ref{eq1}, \ref{eq2} and \ref{eq3} respectively.

After training the EmbedNet, we generate $\phi_i$ and $\psi_i^j$ for word images in the test set and pass them through the EmbedNet to generate $\Phi_i$ and $\Psi_i^j$; these updated embeddings help in selecting the correct predictions with much higher confidence.
% After training the EmbedNet, we pass $\phi_i$ and $\psi_i^j$ through it to generate $\Phi_i$ and $\Psi_i^j$; these updated embeddings help select the correct predictions with much higher confidence.
The reason behind this is, in the updated embedding space the correct text's embeddings lie closer to the input word image's embedding, and the wrong text's embeddings lie farther away from the input word image's embedding.
% As in this case, the correct text's embeddings lie closer to the input word image's embedding, and the wrong text's embeddings lie farther away from the input word image's embedding.
For predicting the text in a given word image $i$, we query $\Psi_i^j \; \forall j \in 1, \dots, K$ using $\Phi_i$ to generate a ranked list of predictions in increasing order of Euclidean distance.
% For calculating the word recognition accuracy, we generate a ranked list of predictions in increasing order of Euclidean distance by querying $\Psi_i^j$ using $\Phi_i$ for a given value of $i$ and $j \in 1, \dots K$.
We consider the word with the least Euclidean distance as the new prediction.

\subsection{Confidence based Accuracy Booster (CAB)} \label{cab}
\begin{figure}[htpb]
    \centerline{\includegraphics[width=0.5\textwidth]{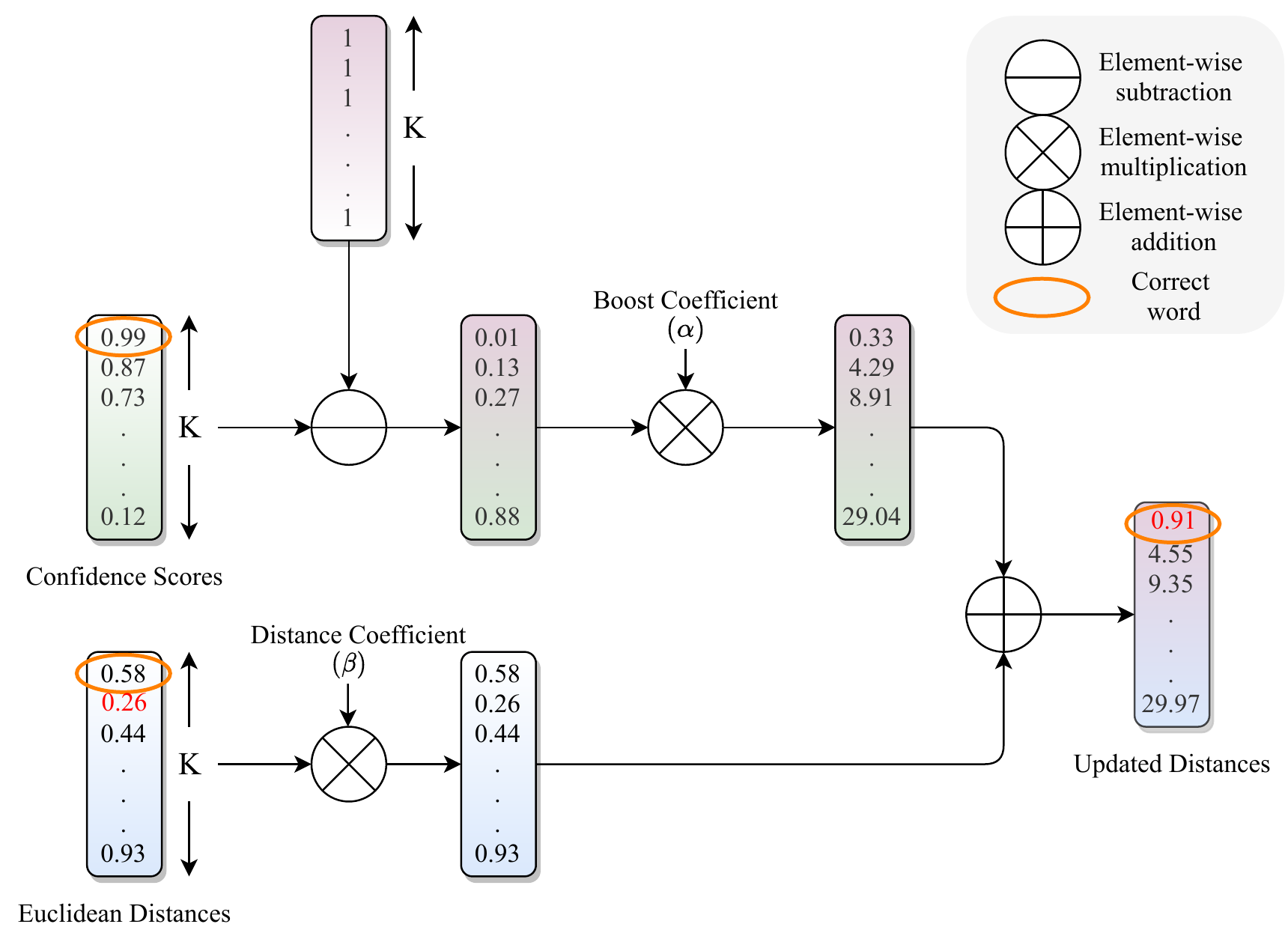}}
    \caption{Confidence based Accuracy Booster (\textsc{cab}) takes in the confidence scores and the Euclidean distances and generates the updated distances.
    The number shown in red indicates the word with the lowest Euclidean distance with $\phi_i$.
    The number circled in orange shows the correct word, which ideally should have the lowest Euclidean distance.
    According to the original Euclidean distances, the correct word is at position 2,
    whereas, the correct word is actually at position 1 (circled in orange).
    At position 1 in the confidence scores vector, we have a value of $0.99$.
    After using \textsc{cab} we get updated distances where the distance for the word at position 1 is the least.
    Therefore, \textsc{cab} helps in incorporating the confidence scores and getting reliable distance values.
    Diagram best viewed in color.
% See text for notation.
    }
    \label{fig:confidence_module}
\end{figure}{}

As shown in Fig. \ref{fig:confidence_module}, 
\textsc{cab} uses a vector of length $K$ consisting of confidence scores, which are a measure of confidence of the \textsc{crnn} for that particular prediction.
% Fig. \ref{fig:confidence_module} shows \textsc{cab}; it uses a vector of length $K$ consisting of the confidence scores; it is a measure of confidence of the \textsc{crnn} for that particular prediction.
Authors in \cite{sid_das} sum this confidence score with the Euclidean distances to improve the word recognition accuracy.
We improve on it and introduce a novel Confidence based Accuracy Booster (\textsc{cab}) as a plug-and-play module.
% which takes in the confidence scores generated from the \textsc{ocr} and the Euclidean distances between $\phi_i$ and $\psi_i^j$.
% It leads to an updated list of distances where words with higher confidence scores have a lower distance value with $\phi_i$.
Mathematically, \textsc{cab} can be defined as:
% Let $f_{cab}$ be the function of \textsc{cab}, defined as:
\begin{equation}
    f_{cab}(\overrightarrow{c}, \overrightarrow{e_d}) = |\overrightarrow{1}-\overrightarrow{c}| \times \alpha \oplus \beta \times \overrightarrow{e_d}
    \label{cab_equ}
\end{equation}
here, $f_{cab}$ is the function for \textsc{cab}, $\overrightarrow{c}$ denotes the vector of confidence scores of length $K$, $\overrightarrow{e_d}$ denotes the vector of Euclidean distances of length $K$, $\overrightarrow{1}$ denotes the vector of ones of length $K$, $\alpha$ is the boost coefficient, $\beta$ is the distance coefficient, and $\oplus$ is the element wise addition operation.
$\alpha$ and $\beta$ are fixed to a constant value. 
$f_{cab}$ takes in $\overrightarrow{c}$ and $\overrightarrow{e_d}$ and generates an updated list of distance where embeddings of words with higher confidence scores have a smaller distance value from $\phi_i$.
% Values of $\alpha$ and $\beta$ 
Using \textsc{cab}, we achieve the highest word recognition accuracy on the validation set when we set the value of $\alpha$ and $\beta$ equal to $33$ and $1$, respectively.
Therefore, we fix the value of $\alpha$ and $\beta$ to $33$ and $1$, unless otherwise stated.
% The values of $\alpha$ and $\beta$ are fixed to $33$ and $1$, respectively, unless otherwise stated.
% Fig. \ref{fig:confidence_module} shows an element-wise subtraction with one on the confidence scores followed by multiplied with a boost coefficient ($\alpha$).
% On the other hand, we multiply the Euclidean distances by the distance coefficient ($\beta$).
% The final updated distance is obtained by summing up the two vectors generated; it has a lower value for the words with higher confidence.

A primary motivation behind creating and using \textsc{cab} is to incorporate the confidence scores generated by the \textsc{crnn}.
As the value of $K$ increases, noise in the predictions increases, which leads to lower confidence score values.
Thus, by updating the distance values using the confidence scores, we can filter out the noisy predictions and select more relevant predictions.

\section{Experiments and Results} \label{experiments}

\subsection{Dataset and evaluation metric details} \label{dataset_details}
\begin{table}[ht!]
    \caption{The dataset consists of pages from the books in the Hindi language.
    The pages are annotated at word-level.
    The annotated words are further divided into train, test, and validation sets.
    % The dataset consists of annotated pages from the books in the Hindi language.
    % The books range from different periods and contain some degraded pages.
}
    \centering
    \begin{tabular}{|c|c|c|c|c|c|}
        \hline
         Language & Annotated & \# Pages & \multicolumn{3}{c|}{\# Word Images}\\
         \hline
         \multirow{2}{*}{Hindi} & \multirow{2}{*}{Yes} & \multirow{2}{*}{$402$}& Train & Validation & Test \\
         \cline{4-6}
          & & & $72,000$ & $8,000$ & $25,475$ \\
         \hline
    \end{tabular}
    \label{tab:dataset}
\end{table}{}

% \begin{itemize}
    % \item Books scanned and annotated internally
    % \item Books range from different time periods sampled from dli
    % \item Add evaluation metric (word accuracy's formula)
    % \item Talk about the split for training MLP and EmbedNet
    % \item Talk about the triplet counts
    % \item Mention why easy negatives are not used for training
    % \item Mention that number of hard, semi-hard and easy negatives change as we processed with the training (as the network learns). It might increase initially (as the network is randomly initialised) and then as the network learns it decreases. 
    % \item As the training progresses the number of hard and semi-hard negatives decreases and easy negatives increases.
%     \item Mention the Euclidean distances between anchor and hard/semi-hard triplets
% \end{itemize}

We perform all the experiments on books in the Hindi language, sampled and annotated from the \textsc{dli} \cite{dli} collection.
These books range from different periods and consist of a variety of font, font sizes, and a few degraded pages.
As summarised in Table \ref{tab:dataset}, the sampled books consist of $402$ pages containing $1,05,475$ words.
We further divide these words into train, validation, and test sets for training and testing the EmbedNet.
We use a pre-trained word recognizer (\textsc{crnn} \cite{stn_crnn}) and an End2End (`\textit{E2E}') network for all the experiments.
We report the word recognition accuracy (\textsc{wra}) for all the experiments performed.
\textsc{wra} is defined as
\begin{equation}
    WRA = \frac{n_r}{n} \times 100,
\end{equation}
where $n_r$ represents the number of correctly recognised words, and $n$ is the total number of words.
\textsc{wra} for methods using $K$ hypotheses is calculated after generating the re-ranked list of predictions.
This list is arranged in the increasing order of Euclidean distance with respect to the query.
The word at the first position of the list is used for calculating the \textsc{wra}.

\begin{table}[ht!]
    \centering
    \caption{Summary of the number of triplet pairs generated for different values of $\gamma$; we further classify them into hard, semi-hard, and easy negatives. 
    We use hard and semi-hard negatives for training the EmbedNet. 
    The number of hard, semi-hard, and easy negatives changes as the training progresses.}
    \begin{tabular}{|c|c|c|c|}
        \hline
        \multirow{3}{*}{$\gamma$} &
        \multicolumn{3}{|c|}{\# triplets} \\
        \cline{2-4}
        &\multicolumn{3}{|c|}{$15,71,820$} \\
        \cline{2-4}
         & \# hard negatives & \# semi-hard negatives & \# easy negatives \\
         \hline
         $0.2$ & $7,894$ & $1,04,468$ & $14,59,458$ \\ \hline
          $0.4$ & $7,894$ & $4,24,590$ & $11,39,336$ \\ \hline
        %   $0.6$ & $7,894$ & $8,88,647$ & $6,75,279$ \\ \hline
        %  $0.7$ & $7,894$ & $10,71,436$ & $4,92,490$ \\ \hline
         $0.8$ & $7,894$ & $11,52,568$ & $4,11,358$ \\ \hline
         $1$ & $7,894$ & $11,64,065$ & $3,99,861$ \\ \hline
    \end{tabular}
    \label{tab:triplets}
\end{table}

As described in Section \ref{EmbedNet_intro}, we generate the triplets and categorize them for training the EmbedNet.
Table \ref{tab:triplets} summarises the number of hard, semi-hard, and easy negatives for different margins ($\gamma$).
With an increase in margin, we observe an increase in the number of semi-hard negatives and a decrease in easy negatives.
It is beneficial to have a larger value for $\gamma$, as it maximizes the number of hard negative and semi-hard negative samples. 
Easy negatives do not contribute to EmbedNet's training, so we do not use them for training the network.
While training the EmbedNet, the number of hard, semi-hard, and easy negatives changes.
Initially, as the network starts from a random initialization, the number of easy negatives is the least while the number of hard and semi-hard negatives are more.
As the training progresses, the number of easy negatives starts to increase while the other two categories decrease.

\subsection{Selection of the best value for the margin}

\begin{table}[!htb]
    \centering
    \caption{EmbedNet's performance for various values of $\gamma$.
    % Values in the `layer size' column denote the input size of a particular layer $l$.
    % The output size of $l$ is equal to the input size of $l + 1$. 
    % The output size of the last layer is always $128$.
    % We select the architecture at \textbf{Sr. No.} \bm{$4$} for performing all the remaining experiments.
    % Accuracies in \textbf{bold} in the `Highest \textsc{wra}' column indicate the best performing model for a given value of $\gamma$.
    }
    \begin{tabular}{|c|c|P {2.5cm}|}
        \hline
        Sr. No. & $\gamma$ & Highest \textsc{wra} (at $K$) \\
        \hline
        1. & $0.2$ & $83.116$ ($2$)\\ \hline
        % 2. & & $2048, 1024, 512$ & \bm{$83.155$} ($2$) \\ \cline{1-1}\cline{3-4}
        % 3. & & $2048, 1024, 512, 256$ & $83.073$ ($2$) \\ \cline{1-1}\cline{3-4}
        % 4. & & $2048, 1024, 512, 256, 128$ & $82.947$ ($2$)\\ \hline
        % 9. & \multirow{4}{*}{$0.7$} & $2048, 1024$ & $83.141$ ($2$) \\ \cline{1-1}\cline{3-4}
        % 9. & & $2048, 1024, 512$ & $83.133$ ($2$) \\ \cline{1-1}\cline{3-4}
        % 9. & & $2048, 1024, 512, 256$ & $82.98$ ($2$) \\ \cline{1-1}\cline{3-4}
        % 9. &  & $2048, 1024, 512, 256, 128$ & $81.884$ ($2$) \\ \hline
        2. & $0.4$ &$83.116$ ($2$)\\ \hline
        % 6. & & $2048, 1024, 512$ & $83.085$ ($2$) \\ \cline{1-1}\cline{3-4}
        % 7. & & $2048, 1024, 512, 256$ & $82.967$ ($2$) \\ \cline{1-1}\cline{3-4}
        % 8. & & $2048, 1024, 512, 256, 128$ & $82.669$ ($2$) \\ \hline
        3. & $0.8$ & $83.226$ ($2$) \\ \hline
        % 10. & & $2048, 1024, 512$ & $83.155$ ($2$)  \\ \cline{1-1}\cline{3-4}
        % 11. & & $2048, 1024, 512, 256$ & $83.061$ ($2$) \\ \cline{1-1}\cline{3-4}
        % 12. & & $2048, 1024, 512, 256, 128$ & $81.915$ ($2$) \\ \hline        
        \textbf{4.} & \bm{$1$} & \bm{$83.242$} (\bm{$2$}) \\ \hline
        % 14. & & $2048, 1024, 512$ & $83.03$ ($2$) \\ \cline{1-1}\cline{3-4}
        % 15. & & $2048, 1024, 512, 256$ & $82.751$ ($2$) \\ \cline{1-1}\cline{3-4}
        % 16. & & $2048, 1024, 512, 256, 128$ & $81.566$ ($2$) \\ \hline
    \end{tabular}
    \label{tab:EmbedNet_comparison}
\end{table}

We train and validate multiple EmbedNets for different $\gamma$ using the train and validation set defined in Table \ref{tab:dataset}.
The aim here is to select the best value of $\gamma$.
For that, we perform the experiments on four different values of $\gamma$.
The results are reported in Table \ref{tab:EmbedNet_comparison}.
% Table \ref{tab:EmbedNet_comparison} reports \textsc{wra} and compares all the models trained using different values of $\gamma$.
EmbedNet with $\gamma$ equal to $1$ has the highest \textsc{wra} on the validation set as compared to a lower values of $\gamma$.
% , and we achieve the highest accuracy on the validation set when $\gamma = 1$, and the number of hidden layers is equal to one.
The reason for this is that a small value of $\gamma$ has a low count of semi-hard negatives (Table \ref{tab:triplets}), which results in reduced triplets for training the EmbedNet.
For the rest of the paper, we consider the value of $\gamma$ equal to $1$ unless otherwise stated.

\subsection{Results and Comparison with various methods} \label{baselines}
% This section presents baseline methods used for assessing the improvement gain after using the EmbedNet with and without \textsc{cab}. 
This section presents baseline methods used for assessing the improvement after using the EmbedNet with and without \textsc{cab}.
We also compare the \textsc{wra} between the baselines and the methods proposed in this work.
% All the results in Table \ref{tab:baselines} are on the test set defined in Table \ref{tab:dataset}.
% The last column in Table \ref{tab:baselines} shows at what value of $K$ (denoted as $K_{high}$) we obtain the highest \textsc{wra} out of the total value of $K$ (shown in ()).
Baseline methods are:
\paragraph{Open-source OCR}
For the first baseline, we use a pre-trained open-source word recognizer: Tesseract \cite{Tesseract}.
The motive here is to compare with an \textsc{ocr} which is not trained on noisy document images.
% The first baseline score shows an open-source \textsc{ocr}'s \cite{Tesseract} performance on the test data.

\paragraph{CRNN} The second baseline score shows the performance of the \textsc{crnn} \cite{stn_crnn} trained using the best path decoding algorithm.
It was trained on $Dataset1$ defined in \cite{sid_das}.
Here, we generate a single prediction for each test image.

% \paragraph{E2E w/o C} We generate the third baseline score using the method proposed in \cite{sid_das}; for that, we use the embeddings generated from \textit{E2E} and multiple ($K$) hypotheses generated from the \textsc{crnn}.
% For calculating the \textsc{wra} of a given word image $i$, we perform a nearest neighbor's search on $\psi_i^j \; \forall j \in 1, \dots, K$ using $\phi_i$.
% We refer to this method as `\textit{E2E w/o C}'.
% Table \ref{tab:baselines} shows the \textsc{wra} generated using this method, we get a \textsc{wra} of $82.685$ at $K=2$, and for higher values $K$, the \textsc{wra} decreases.

\paragraph{E2E+C} We generate the third baseline score using the method proposed in \cite{sid_das}; for that, we use the embeddings generated from \textit{E2E} and multiple ($K$) hypotheses generated from the \textsc{crnn}. 
% We generate a new score by adding the confidence information to the Euclidean distances.
% We follow the same method as \textit{E2E w/o C}; the only difference is that we add the confidence information to the distances obtained after the nearest neighbor's search.
For calculating the \textsc{wra} of a given word image $i$, we perform a nearest neighbor's search on $\psi_i^j \; \forall j \in 1, \dots, K$ using $\phi_i$ and add the confidence information to the distances obtained after the nearest neighbor's search; this provides us a re-ranked list, from which we consider the word with the least Euclidean distance as the new prediction.
We refer to this method as `\textit{E2E+C}'.
% Using \textit{E2E+C}, we obtain a \textsc{wra} of 83.062 at $K=2$.
% Similar to the observation for \textit{E2E}, the \textsc{wra} decreases for the higher values of $K$.

\begin{figure*}
    \centering
    \includegraphics[width=\textwidth]{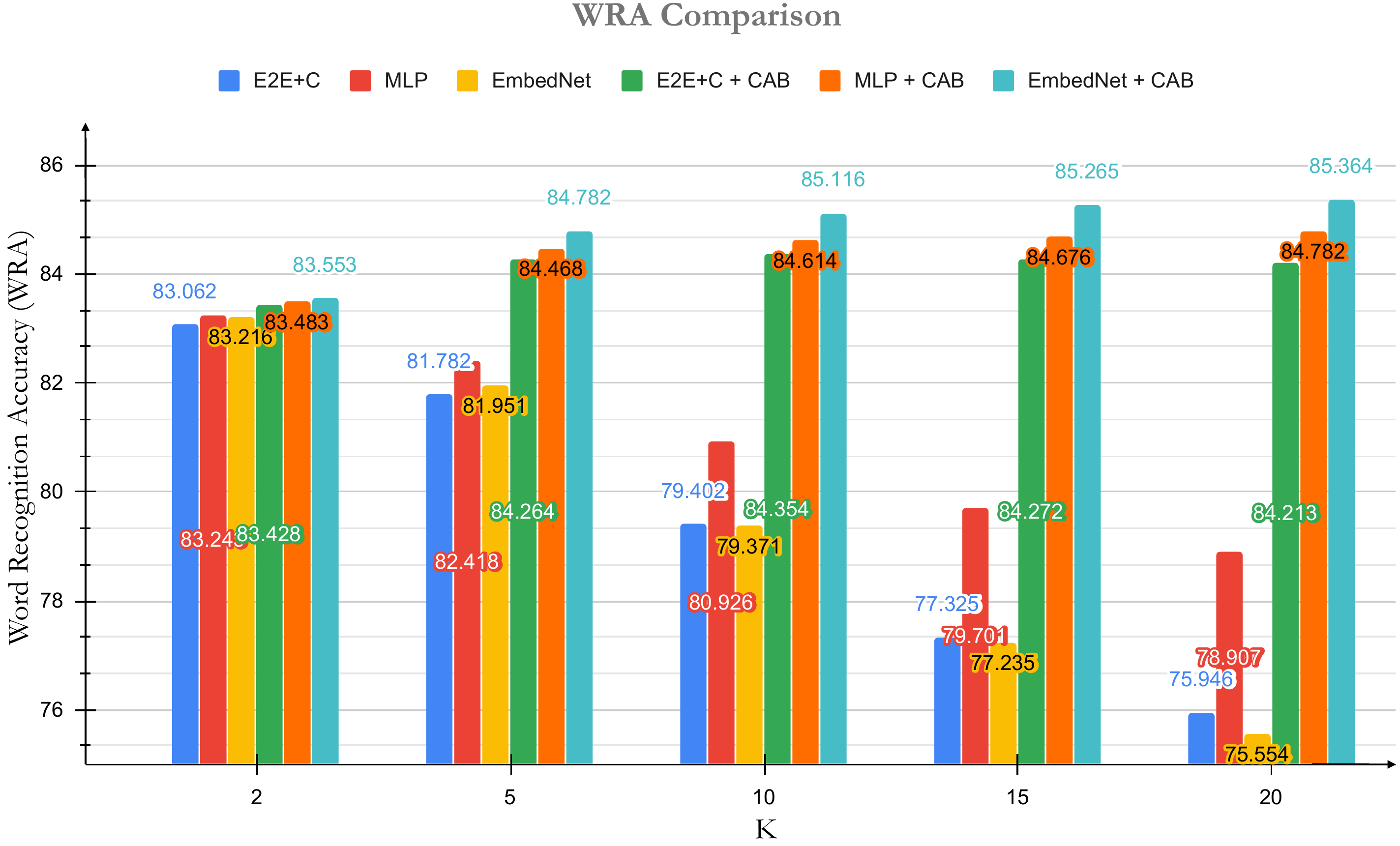}
    \caption{Comparison between the \textsc{wra} for \textit{E2E+C}, \textsc{mlp}, and EmbedNet with and without \textsc{cab}.
For the experiments not using \textsc{cab}, the \textsc{wra} first increases and then starts to decrease.
The reason for such a trend is, as $K$ increases, the noise in the \textsc{crnn}'s predictions increases leading to lower \textsc{wra}.
However, using \textsc{cab} helps avoid this issue, as it uses the confidence scores from the \textsc{crnn}, which decreases as the noise increases.
We achieve the highest \textsc{wra} of $85.364$ using EmbedNet + \textsc{cab} at $K=20$.}
    \label{fig:wra_comparison}
\end{figure*}

\paragraph{Multilayared Perceptron} For calculating the last baseline score, we train a Multi-Layered Perceptron (\textsc{mlp}) on the train data defined in Table \ref{tab:dataset}.
\textsc{mlp} is a function $f_{mlp}$ defined as $f_{mlp}(\phi_i) = \hat{\phi_i}$; it projects $\phi_i$ to an updated embedding space where the Euclidean distance between $\hat{\phi_i}$ and correct $\psi_i^j$ is less than the distance between $\phi_i$ and correct $\psi_i^j$.
\textsc{mlp} consists of three layers, the initial and final layers have the input and output dimensions of $2048$, respectively; the hidden layer has the input and output dimensions of $256$ and $128$, respectively.
The ReLU activation function follows each layer to introduce non-linearity.
Mean Squared Error (\textsc{mse}) is used as a loss function for training the \textsc{mlp}.
We train the network for $150$ epochs with early stopping acting as a regularizer and use the Adam optimizer with a constant learning rate of $0.0001$.
For calculating the word recognition accuracy, we query $\psi_i^j$ using $\hat{\phi_i}$ for a given value of $i$ and $j \in 1, \dots, K$ to get a ranked list of predictions in increasing order of Euclidean distance.
We consider the word with minimum Euclidean distance as the new prediction.

\begin{table}[!htb]
    \centering
    \caption{Comparison between the \textsc{wra} of all the baselines and the methods proposed in this work.
$K_{high}$ signifies $K$'s value at which we achieve the highest \textsc{wra}; ($K$) signifies the maximum value of $K$ for that experiment.}
    \begin{tabular}{|c|c|c|c|}
        \hline
         Sr. No. & Method & \textsc{wra} & $K_{high} (K)$ \\
         \hline
         1. & Tesseract \cite{Tesseract} & $35.435$ & $1$ $(1)$ \\
         \hline
         2. & \textsc{crnn} \cite{stn_crnn} & $81.543$ & $1$ $(1)$ \\
         \hline
        %  3. & \textit{E2E w/o C} \cite{sid_das} & $82.685$ & $3$ $(20)$ \\
        %  \hline
         3. & \textit{E2E+C} \cite{sid_das} & $83.062$ & $2$ $(20)$ \\
         \hline
         4. & \textit{E2E+C} + \textsc{cab} (ours) & $84.358$ & $11$ $(20)$ \\
         \hline
        %  4. & \textit{E2E+C + \textsc{cab}} \cite{sid_das} & $83.062$ & $2$ $(20)$ \\
        %  \hline
         5. & \textsc{mlp} (ours) & $83.259$ & $3$ $(20)$\\
         \hline
         6. & EmbedNet (ours) & $83.216$ & $2$ $(20)$ \\
         \hline
         7. & \textsc{mlp + cab} (ours) & $84.782$ & $20$ $(20)$\\
         \hline
         8. & \textbf{EmbedNet + \textsc{cab} (ours)} & \bm{$85.364$} & \bm{$20$} \bm{$(20)$} \\
         \hline
    \end{tabular}
    \label{tab:baselines}
\end{table}

Table \ref{tab:baselines} contrasts the \textsc{wra} of all the baselines with the methods proposed in this work.
We observe the lowest \textsc{wra} for the methods not using multiple hypotheses, i.e., methods for which we have a maximum value of $K$ equal to $1$.
Using~\cite{Tesseract}, we achieve a \textsc{wra} of $35.435$ on the test set.
As the training of the open-source \textsc{ocr} does not take place on the noisy documents that we are using, it performs the worst; this shows that the data that we are using contains highly degraded word images which are difficult to understand.
On the other hand, we train a \textsc{crnn} \cite{stn_crnn} on the train split defined in Table \ref{tab:dataset} and achieve a \textsc{wra} of $81.543$ on the test set.

We observe an improved \textsc{wra} for the methods using multiple hypotheses, i.e., methods for which we have a maximum value of $K$ higher than $1$; in all the experiments, we have the maximum value of $K$ equal to $20$.
We observe the \textsc{wra} plateauing on the validation data for higher values of $K$; due to this, we choose to limit the highest value of $K$ at $20$.
\textit{E2E+C} achieves the maximum \textsc{wra} of $83.062$.
However, as we observe in Fig. \ref{fig:wra_comparison}, it achieves a maximum \textsc{wra} at a small value of $K$ ($2$); \textsc{wra} begins to decrease as we increase $K$.
So, when using \textit{E2E+C}, one cannot use $K$ higher than two, making it impractical to use.
We add \textsc{cab} to \textit{E2E+C} and observe a performance gain and more consistent \textsc{wra} values for a higher value of $K$.
Using \textit{E2E+C} + \textsc{cab}, we achieve the highest \textsc{wra} of $84.358$ at $K=11$.
Fig. \ref{fig:wra_comparison} shows the change in \textsc{wra} on increasing the value of $K$; we observe a steady increase till $K=11$, after which the \textsc{wra} starts to decrease.
However, this decrease in the \textsc{wra} is very small as compared to \textit{E2E+C} without \textsc{cab}.
Even at $K=20$, \textit{E2E+C} + \textsc{cab} achieves $8.267$ more \textsc{wra} as compared to \textit{E2E+C} at $K=20$.
The reason for such stability is the usage of the confidence scores.
As $K$ increases, the noise present in the \textsc{ocr}'s predictions also increases, leading to a lower confidence score for the noisy predictions.
\textsc{cab} uses this fact and results in better and consistent \textsc{wra}.

We observe an improvement in the \textsc{wra} for \textsc{mlp} and EmbedNet without \textsc{cab} as compared to \textit{E2E+C}, \textsc{crnn}, and Tesseract~\cite{Tesseract}.
\textsc{mlp} and EmbedNet achieve the highest \textsc{wra} of $83.259$ and $83.216$, respectively.
As observed in the case of \textit{E2E+C} and shown in Fig. \ref{fig:wra_comparison}, the \textsc{wra} starts to decrease as the value of $K$ is increased, making them impractical to use.
Upon using \textsc{cab} with \textsc{mlp} and EmbedNet, we observe high gains in the \textsc{wra} for large $K$ as shown in Fig. \ref{fig:wra_comparison}.
\textsc{mlp} + \textsc{cab} attains the highest \textsc{wra} at $K=20$, equal to $84.782$, which is $8.836$ more than the \textsc{mlp} without \textsc{cab} at $K=20$.
As pointed out in Section \ref{EmbedNet_intro}, EmbedNet not only helps in bringing correct $\Psi_i^j$ closer to $\Phi_i$ but also pushes incorrect $\Psi_i^j$ farther away from $\Phi_i$.
\textsc{cab} utilizes this fact, and as we can see, we obtain a \textsc{wra} of $85.364$, which is $9.418$ more than the \textit{E2E+C} without \textsc{cab} at $K=20$.

Hence, by using the \textsc{cab}, we observe substantial gains in \textsc{wra} as $K$ increases, and we also see a more steady \textsc{wra} for all the values of $K$; this enables us to freely choose any value of $K$ without any loss of \textsc{wra}, which was not possible while using \textit{E2E+C}, \textsc{mlp}, and EmbedNet without \textsc{cab}.

\subsection{Qualitative Results} \label{Qualitative_results}
\begin{figure}[!htpb]
    \centering
    \includegraphics[width=0.48\textwidth]{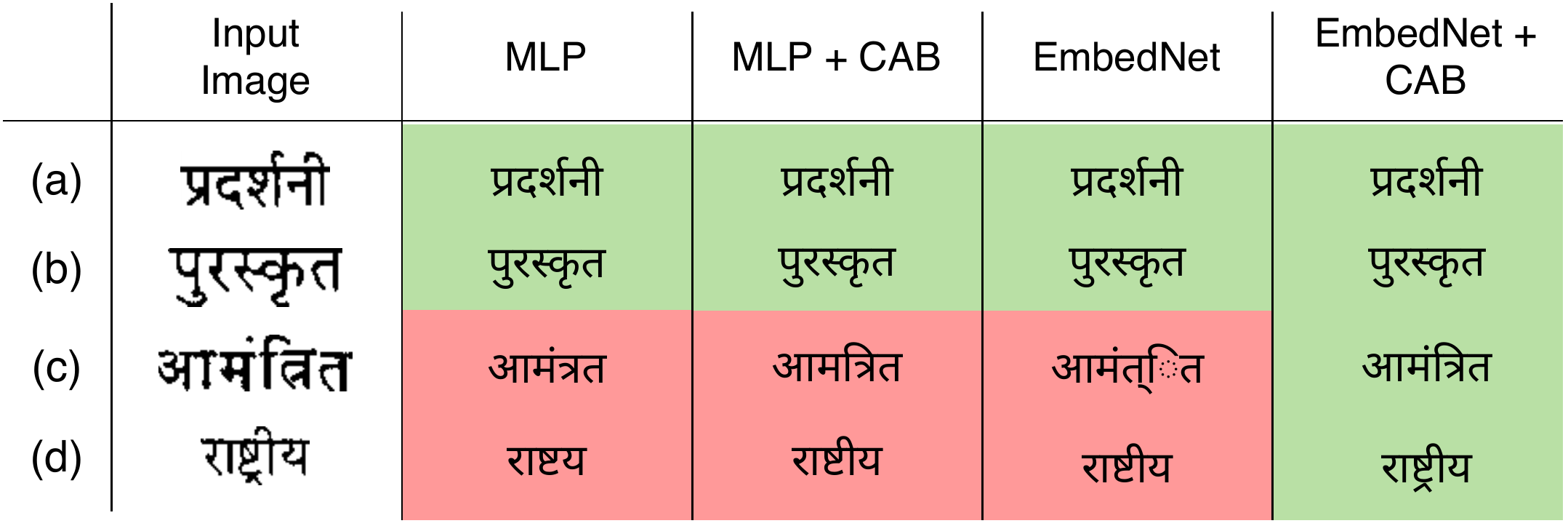}
    \caption{Qualitative results on randomly chosen word images after processing using \textsc{mlp} and EmbedNet with and without \textsc{cab}.
    Diagram best viewed in color.}
    \label{fig:quantitative}
\end{figure}

Fig. \ref{fig:quantitative} shows qualitative results on some randomly chosen words.
Words in Fig. \ref{fig:quantitative} (a) and (b) are long and contain characters and contain half consonants.
Both of the words are recognised perfectly by \textsc{mlp} and WordNet with and without \textsc{cab}.
However, words in Fig. \ref{fig:quantitative} (c) and (d) contains rare characters and are distorted, due to this \textsc{mlp} with and without \textsc{cab}, and EmbedNet fail to predict the correct word.
EmbedNet + \textsc{cab} performs well for these cases and is able to predict the correct word.
This shows the ability of EmbedNet to use the complementary information provided by word recognition and word image embedding methods.
% This shows the ability of EmbedNet to use the complementary information provided by recognition-based and recognition-free methods.

\subsection{Computational Costs}
% \begin{itemize}
%     \item Discuss the time table
% \end{itemize}
\begin{table}[!htb]
    \centering
    \caption{Time taken by various processes in the word recognition pipeline.
    Time for the methods dependent on $K$ is calculated for $K=20$.
    Values are reported for a single word's image/text.}
    \begin{tabular}{|P {0.8cm}|P {2.8cm}|P {2cm}|P {1.5cm}|}
         \hline
        Mode & Process & Average time (in milliseconds) & Dependent on \\ \hline
        \multirow{3}{*}{Offline\vspace{-0.59cm}} & Text from \textsc{crnn} & $580$ & $K$ \\ \cline{2-4}
        & Text embeddings' generation & $18.2$ & $K$ \\ \cline{2-4}
        & Image embeddings' generation & $23.89$ & $K$ \\ \hline
        \multirow{4}{*}{Online\vspace{-0.59cm}} & EmbedNet pass & $0.21$ & Network's size \\ \cline{2-4}
        & \textsc{mlp} pass & $0.22$ & Network's size \\ \cline{2-4}
        & \textsc{wra} calculation & $0.24$ & $K$ \\ \cline{2-4}
        & \textsc{wra} calculation with \textsc{cab} & $0.27$ & $K$ \\ \hline
    \end{tabular}
    \label{tab:time}
\end{table}

Table \ref{tab:time} shows the time taken for various processes done in the entire pipeline.
All the experiments are performed on  Intel Xeon E5-2640 v4 processors with $32$ \textsc{gb} \textsc{ram} on \textsc{nvidia geforce} \textsc{gtx} 1080 Ti \textsc{gpu}.
For calculating the time taken, we run the experiments $10$ times and average the time taken in all the runs.
The process of calculating the word accuracies for all the values of $K$ is parallelizable; this reduces the time taken by $\frac{1}{K}$.

There are two modes in which the majority of our experiments take place.
The first mode is the offline mode, which involves computations required only once.
It includes time taken in generating the \textsc{ocr} output for all the $n$ word images and the time taken by the End2End network in generating $\phi_i$ and $\psi_i^j$.
Second is the online mode, which includes computations that are required every time we calculate \textsc{wra}.
It includes time taken in passing the embeddings through the \textsc{mlp} and EmbedNet.
It also includes the time taken in calculating \textsc{wra} with and without \textsc{cab}.

\section{Conclusion} \label{conclusion}
To summarise, in this work, we aim at fusing the word recognition and word image embedding approaches for word recognition.
% To summarise, in this work, we aim at fusing the recognition-based and recognition-free approaches for word recognition.
For achieving this, we propose EmbedNet, which helps in learning an updated Euclidean space.
We also propose \textsc{cab} for using the updated Euclidean space and boosting the \textsc{wra} by approximately 10\% at $K=20$.
We show that learning based approaches for fusion show more promising results than rule-based fusion.
As a future task, we plan to develop an end-to-end architecture capable of fusing word recognition and word image embedding approaches.
% As a future task, we plan to develop an end-to-end architecture capable of fusing recognition-based and recognition-free approaches.

\bibliographystyle{IEEEtran}
\bibliography{ref}

\begin{thebibliography}{10}
\providecommand{\url}[1]{#1}
\csname url@samestyle\endcsname
\providecommand{\newblock}{\relax}
\providecommand{\bibinfo}[2]{#2}
\providecommand{\BIBentrySTDinterwordspacing}{\spaceskip=0pt\relax}
\providecommand{\BIBentryALTinterwordstretchfactor}{4}
\providecommand{\BIBentryALTinterwordspacing}{\spaceskip=\fontdimen2\font plus
\BIBentryALTinterwordstretchfactor\fontdimen3\font minus
  \fontdimen4\font\relax}
\providecommand{\BIBforeignlanguage}[2]{{%
\expandafter\ifx\csname l@#1\endcsname\relax
\typeout{** WARNING: IEEEtran.bst: No hyphenation pattern has been}%
\typeout{** loaded for the language `#1'. Using the pattern for}%
\typeout{** the default language instead.}%
\else
\language=\csname l@#1\endcsname
\fi
#2}}
\providecommand{\BIBdecl}{\relax}
\BIBdecl

\bibitem{why_hindi}
K.~Dutta, P.~Krishnan, M.~Mathew, and C.~V. Jawahar, ``{Towards Accurate
  Handwritten Word Recognition for Hindi and Bangla},'' in \emph{Computer
  Vision, Pattern Recognition, Image Processing, and Graphics}, 2018.

\bibitem{CRNN_begins}
B.~{Shi}, X.~{Bai}, and C.~{Yao}, ``{An End-to-End Trainable Neural Network for
  Image-Based Sequence Recognition and Its Application to Scene Text
  Recognition},'' \emph{IEEE Transactions on Pattern Analysis and Machine
  Intelligence}, 2017.

\bibitem{Chen}
Z.~{Chen}, Y.~{Wu}, F.~{Yin}, and C.~{Liu}, ``Simultaneous script
  identification and handwriting recognition via multi-task learning of
  recurrent neural networks,'' in \emph{International Conference on Document
  Analysis and Recognition (ICDAR)}, 2017.

\bibitem{Sun}
Z.~{Sun}, L.~{Jin}, Z.~{Xie}, Z.~{Feng}, and S.~{Zhang}, ``Convolutional
  multi-directional recurrent network for offline handwritten text
  recognition,'' in \emph{Conference on Frontiers in Handwriting Recognition
  (ICHFR)}, 2016.

\bibitem{Garain}
U.~{Garain}, L.~{Mioulet}, B.~B. {Chaudhuri}, C.~{Chatelain}, and T.~{Paquet},
  ``{Unconstrained Bengali handwriting recognition with recurrent models},'' in
  \emph{International Conference on Document Analysis and Recognition (ICDAR)},
  2015.

\bibitem{Adak}
C.~{Adak}, B.~B. {Chaudhuri}, and M.~{Blumenstein}, ``{Offline Cursive Bengali
  Word Recognition Using CNNs with a Recurrent Model},'' in \emph{International
  Conference on Frontiers in Handwriting Recognition (ICHFR)}, 2016.

\bibitem{Kris}
P.~{Krishnan}, K.~{Dutta}, and C.~V. {Jawahar}, ``{Deep Feature Embedding for
  Accurate Recognition and Retrieval of Handwritten Text},'' in
  \emph{International Conference on Frontiers in Handwriting Recognition
  (ICHFR)}, 2016.

\bibitem{Sudholt}
S.~{Sudholt} and G.~A. {Fink}, ``{PHOCNet: A Deep Convolutional Neural Network
  for Word Spotting in Handwritten Documents},'' in \emph{2016 15th
  International Conference on Frontiers in Handwriting Recognition (ICFHR)},
  2016.

\bibitem{Wilkinson}
T.~{Wilkinson} and A.~{Brun}, ``{Semantic and Verbatim Word Spotting Using Deep
  Neural Networks},'' in \emph{International Conference on Frontiers in
  Handwriting Recognition (ICHFR)}, 2016.

\bibitem{Sudhlt}
S.~Sudholt and G.~Fink, ``{Attribute CNNs for Word Spotting in Handwritten
  Documents},'' \emph{International Journal on Document Analysis and
  Recognition (IJDAR)}, 2017.

\bibitem{Tesseract}
R.~Smith, ``{An Overview of the Tesseract OCR Engine},'' in \emph{International
  Conference on Document Analysis and Recognition (ICDAR)}, 2007.

\bibitem{prec_recall}
P.~Krishnan, R.~Shekhar, and C.~Jawahar, ``{Content level access to Digital
  Library of India pages},'' in \emph{ACM International Conference Proceeding
  Series (ICPS)}, 2012.

\bibitem{word_seg}
A.~{Gordo}, J.~{Almazán}, N.~{Murray}, and F.~{Perronin}, ``{LEWIS: Latent
  Embeddings for Word Images and Their Semantics},'' in \emph{International
  Conference on Computer Vision (ICCV)}, 2015.

\bibitem{sid_das}
S.~Bansal, P.~Krishnan, and C.~V. Jawahar, ``{Fused Text Recogniser and Deep
  Embeddings Improve Word Recognition and Retrieval},'' in \emph{Document
  Analysis Systems (DAS)}, 2020.

\bibitem{e2e}
P.~{Krishnan}, K.~{Dutta}, and C.~V. {Jawahar}, ``{Word Spotting and
  Recognition Using Deep Embedding},'' in \emph{IAPR International Workshop on
  Document Analysis Systems (DAS)}, 2018.

\bibitem{ctc}
A.~Graves, S.~Fern\'{a}ndez, F.~Gomez, and J.~Schmidhuber, ``{Connectionist
  Temporal Classification: Labelling Unsegmented Sequence Data with Recurrent
  Neural Networks},'' in \emph{International Conference on Machine Learning
  (ICML)}, 2006.

\bibitem{manmatha96}
R.~Manmatha, C.~Han, and E.~M. Riseman, ``Word spotting: A new approach to
  indexing handwriting,'' in \emph{Computer Vision and Pattern Recognition,
  {CVPR}}, ser. CVPR '96, 1996, p. 631.

\bibitem{Rath}
T.~Rath and R.~Manmatha, ``{Word spotting for historical documents},'' in
  \emph{International Journal of Document Analysis and Recognition (IJDAR)},
  2007.

\bibitem{Balasubramanian}
A.~Balasubramanian, M.~Meshesha, and C.~V. Jawahar, ``Retrieval from document
  image collections,'' in \emph{Document Analysis Systems (DAS)}, 2006.

\bibitem{Shekhar}
R.~{Shekhar} and C.~V. {Jawahar}, ``{Word Image Retrieval Using Bag of Visual
  Words},'' in \emph{Document Analysis Systems (DAS)}, 2012.

\bibitem{AlmazanPAMI14}
J.~Almaz{\'{a}}n, A.~Gordo, A.~Forn{\'{e}}s, and E.~Valveny, ``Word spotting
  and recognition with embedded attributes,'' \emph{PAMI}, 2014.

\bibitem{Jaderberg_2}
M.~Jaderberg, K.~Simonyan, A.~Vedaldi, and A.~Zisserman, ``Synthetic data and
  artificial neural networks for natural scene text recognition,'' in
  \emph{Workshop on Deep Learning, NIPS}, 2014.

\bibitem{Jaderberg_3}
M.~Jaderberg, A.~Vedaldi, and A.~Zisserman, ``{Deep Features for Text
  Spotting},'' in \emph{European Conference on Computer Vision (ECCV)}, 2014.

\bibitem{VGG}
K.~Simonyan and A.~Zisserman, ``Very deep convolutional networks for
  large-scale image recognition,'' in \emph{International Conference on
  Learning Representations}, 2015.

\bibitem{Poznanski}
A.~{Poznanski} and L.~{Wolf}, ``{CNN-N-Gram for Handwriting Word
  Recognition},'' in \emph{Computer Vision and Pattern Recognition (CVPR)},
  2016.

\bibitem{Bhardwaj}
A.~Bhardwaj, S.~Kompalli, S.~Setlur, and V.~Govindaraju, ``{An OCR based
  approach for word spotting in Devanagari documents},'' in \emph{Document
  Recognition and Retrieval Conference (DRR)}, 2008.

\bibitem{Chaudhury}
S.~Chaudhury, G.~Sethi, A.~Vyas, and G.~Harit, ``{Devising interactive access
  techniques for Indian language document images},'' in \emph{International
  Conference on Document Analysis and Recognition (ICDAR).}, 2003.

\bibitem{stn_crnn}
K.~{Dutta}, P.~{Krishnan}, M.~{Mathew}, and C.~V. {Jawahar}, ``{Improving
  CNN-RNN Hybrid Networks for Handwriting Recognition},'' in
  \emph{International Conference on Frontiers in Handwriting Recognition
  (ICFHR)}, 2018.

\bibitem{resnet18}
K.~He, X.~Zhang, S.~Ren, and J.~Sun, ``{Deep Residual Learning for Image
  Recognition},'' \emph{CoRR}, 2015.

\bibitem{facenet}
F.~{Schroff}, D.~{Kalenichenko}, and J.~{Philbin}, ``{FaceNet: A unified
  embedding for face recognition and clustering},'' in \emph{Conference on
  Computer Vision and Pattern Recognition (CVPR)}, 2015.

\bibitem{dli}
V.~Ambati, N.~Balakrishnan, R.~Reddy, L.~Pratha, and C.~V. Jawahar, ``{The
  Digital Library of India Project: Process, Policies and Architecture},'' in
  \emph{Second International Conference on Digital Libraries (ICDL)}, 2007.

\end{thebibliography}
\end{document}